\pdfoutput=1

\documentclass[twocolumn]{article}

\usepackage[preprint]{acl}

\usepackage{times}
\usepackage{pdfpages}
\usepackage{latexsym}
\usepackage{array}
\usepackage{lipsum}  
\usepackage{makecell}
\usepackage{booktabs}
\usepackage{graphicx}
\usepackage{caption}
\usepackage{booktabs}
\usepackage{xcolor}
\usepackage{enumitem}
\usepackage{array}
\usepackage{longtable}
\usepackage{natbib} 
\usepackage{standalone}
\usepackage{subcaption}
\usepackage{tikz}
\usepackage{pgfplots}
\pgfplotsset{compat=1.18}
\usepackage{array}
\usepackage[ruled,vlined]{algorithm2e}
\usepackage{colortbl}  
\usepackage{array}     
\usepackage{amsmath}
\usepackage{booktabs}  
\usepackage[T1]{fontenc}
\SetCommentSty{textit}  
\SetVlineSkip{0.5em}   

\usepackage[utf8]{inputenc}

\usepackage{microtype}

\usepackage{inconsolata}

\usepackage{graphicx}
\usepackage{algpseudocode}
\usepackage{multirow}
\usepackage{tabularx}   
\usepackage{amsmath}
\usepackage[utf8]{inputenc}
\usepackage{verbatim} 
\usepackage{amsthm}
\usepackage{amssymb}
\usepackage{pifont}
\usepackage{fontawesome}

\title{MedHallu: A Comprehensive Benchmark for Detecting Medical Hallucinations in Large Language Models}

\author{
 \textbf{Shrey Pandit\textsuperscript{1}\thanks{Corresponding author: \href{mailto:shreypandit@utexas.edu}{shreypandit@utexas.edu}}},
 \textbf{Jiawei Xu\textsuperscript{1}},
 \textbf{Junyuan Hong\textsuperscript{1}},\\
 \textbf{Zhangyang Wang\textsuperscript{1}},
 \textbf{Tianlong Chen\textsuperscript{2}},
 \textbf{Kaidi Xu\textsuperscript{3}},
 \textbf{Ying Ding\textsuperscript{1}}
\\
\faGlobe~Dataset \& Code: \url{https://medhallu.github.io/} \\
 \textsuperscript{1}University of Texas at Austin,
 \textsuperscript{2}UNC Chapel Hill,
 \textsuperscript{3}Drexel University,
}

\begin{document}
\maketitle
\begin{abstract}
Advancements in Large Language Models (LLMs) and their increasing use in medical question-answering necessitate rigorous evaluation of their reliability. A critical challenge lies in hallucination, where models generate plausible yet factually incorrect outputs. In the medical domain, this poses serious risks to patient safety and clinical decision-making. To address this, we introduce \textbf{MedHallu}, the first benchmark specifically designed for medical hallucination detection. MedHallu comprises 10,000 high-quality question-answer pairs derived from PubMedQA, with hallucinated answers systematically generated through a controlled pipeline. Our experiments show that state-of-the-art LLMs, including GPT-4o, Llama-3.1, and the medically fine-tuned UltraMedical, struggle with this binary hallucination detection task, with the best model achieving an F1 score as low as 0.625 for detecting ``hard'' category hallucinations. Using bidirectional entailment clustering, we show that harder-to-detect hallucinations are semantically closer to ground truth. Through experiments, we also show incorporating domain-specific knowledge and introducing a ``not sure'' category as one of the answer categories improves the precision and F1 scores by up to 38\% relative to baselines. 
\end{abstract}

\section{Introduction}
Recent advances in Large Language Models (LLMs)~\citep{achiam2023gpt} have catalyzed their widespread adoption as assistive tools across a multitude of domains, including software development~\citep{Krishna_2024_software}, healthcare ~\citep{singhal2022largelanguagemodelsencode_health}, weather prediction~\citep{li2024cllmatemultimodalllmweather}, and financial applications \citep{nie2024surveylargelanguagemodels}. However, LLMs are prone to hallucination~\citep{bang2023multitaskmultilingualmultimodalevaluation_hallucination}, where they generate plausible but factually incorrect or unverifiable information~\citep{Ji_2023_12RW, Huang_2025_survey11}. Hallucinations can arise from various factors, including biased or insufficient training data~\citep{han2024skipnsimplemethod_data, zhang2024knowledgeovershadowingcausesamalgamated_data}, and inherent architectural limitations of LLMs \citep{leng2023mitigatingobjecthallucinationslarge_architecture, kalai2024calibratedlanguagemodelshallucinate_architecture}. This issue is particularly problematic in high-stakes fields such as the medical domains, where the generation of incorrect information can exacerbate health disparities~\citep{singhal2022largelanguagemodelsencode_health}. 


\begin{figure}[t]
\centering
\includegraphics[width=1\linewidth]{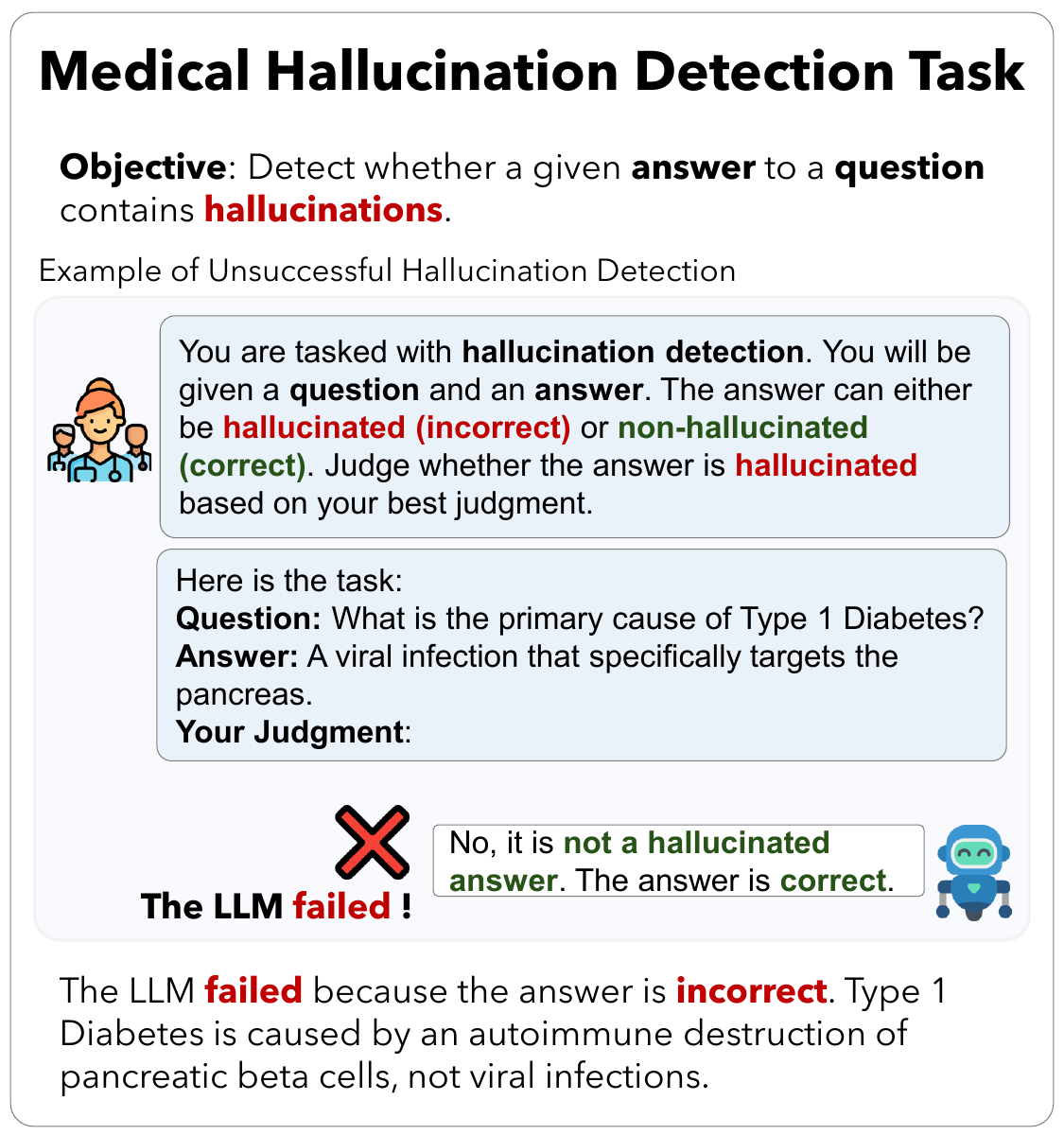}
\caption{An example of medical hallucination detection. The detailed prompt used for the hallucination detection task is presented in Appendix \ref{appendix:prompt}.}
\label{fig:task}
\end{figure}

Detecting hallucinations in LLM outputs (Figure~\ref{fig:task}) is therefore of critical importance. Various methods have been proposed to address this issue, including self-consistency~\citep{wang2023selfconsistencyimproveschainthought}, sampling-based approaches such as SelfCheckGPTZero \citep{manakul2023selfcheckgptzeroresourceblackboxhallucination_detect1}, and intrinsic methods that evaluate token-level uncertainty and entropy~\citep{azaria2023internalstatellmknows_detect2, xiao2021hallucinationpredictiveuncertaintyconditional_detect3}.
Existing benchmarks, such as HaluEval~\citep{Hallueval} and Haydes~\citep{liu2022tokenlevelreferencefreehallucinationdetection} primarily evaluate hallucination detection capabilities on general tasks, including summarization, question answering, and dialogue systems, with an emphasis on common-sense knowledge rather than domain specificity. 
This gap becomes particularly consequential in the medical domains, where specialized terminology requires precise handling, as minor lexical deviations can lead to substantially divergent interpretations~\citep{singhal2022largelanguagemodelsencode_health}. While recent efforts such as HaluBench~\citep{ravi2024lynxopensourcehallucination}, incorporate limited samples from the medical domains, their domain-agnostic generation frameworks lack medical curation. Similarly, Med-Halt~\citep{pal2023medhaltmedicaldomainhallucination} focuses on model benchmarking rather than providing a structured evaluation resource. Furthermore, the subtlety of hallucinations (e.g., whether they are hard or easy to detect) remains underexplored in the medical context. Additionally, the performance differences between pre-trained LLMs and fine-tuned medical LLMs are sparsely documented~\cite{ravi2024lynxopensourcehallucination, Hallueval, pal2023medhaltmedicaldomainhallucination}.

To address these gaps, we present the \textbf{Med}ical \textbf{Hallu}cination detection dataset (\textbf{MedHallu}), a comprehensive corpus of 10,000 medical question-answer pairs derived from the established PubMedQA dataset. Each pair is meticulously annotated to distinguish accurate responses from hallucinated content. Furthermore, MedHallu is stratified into easy, medium, and hard detection tiers based on the subtlety of hallucinations, enabling granular evaluation of model capabilities. The primary contributions of this research are threefold:
 \begin{itemize}
 \vspace{-3mm}
     \item We introduce MedHallu, one of the first datasets specifically designed for medical hallucination detection tasks. Comprising 10,000 entries derived from PubMedQA, MedHallu is systematically categorized into three levels of difficulty—easy, medium, and hard—based on the subtlety of hallucination detection.

     \item We find that hallucinated answers that are semantically closer to the ground truth are more challenging to detect. Furthermore, clustered answers using bi-directional entailment reveal uniformity, where all entries in a cluster are consistently either easy or hard to detect.

    \item Our evaluation shows that general-purpose LLMs outperform fine-tuned medical LLMs in medical hallucination detection tasks. Additionally, we find that model performance can be enhanced by providing relevant knowledge to LLMs. Moreover, introducing a ``not sure'' class alongside the existing classes of ``hallucinated'' and ``not-hallucinated'' leads to improved precision, which is critical in the medical domains.

 \end{itemize}



\begin{figure*}[ht]
\centering
\includegraphics[width=1\linewidth]{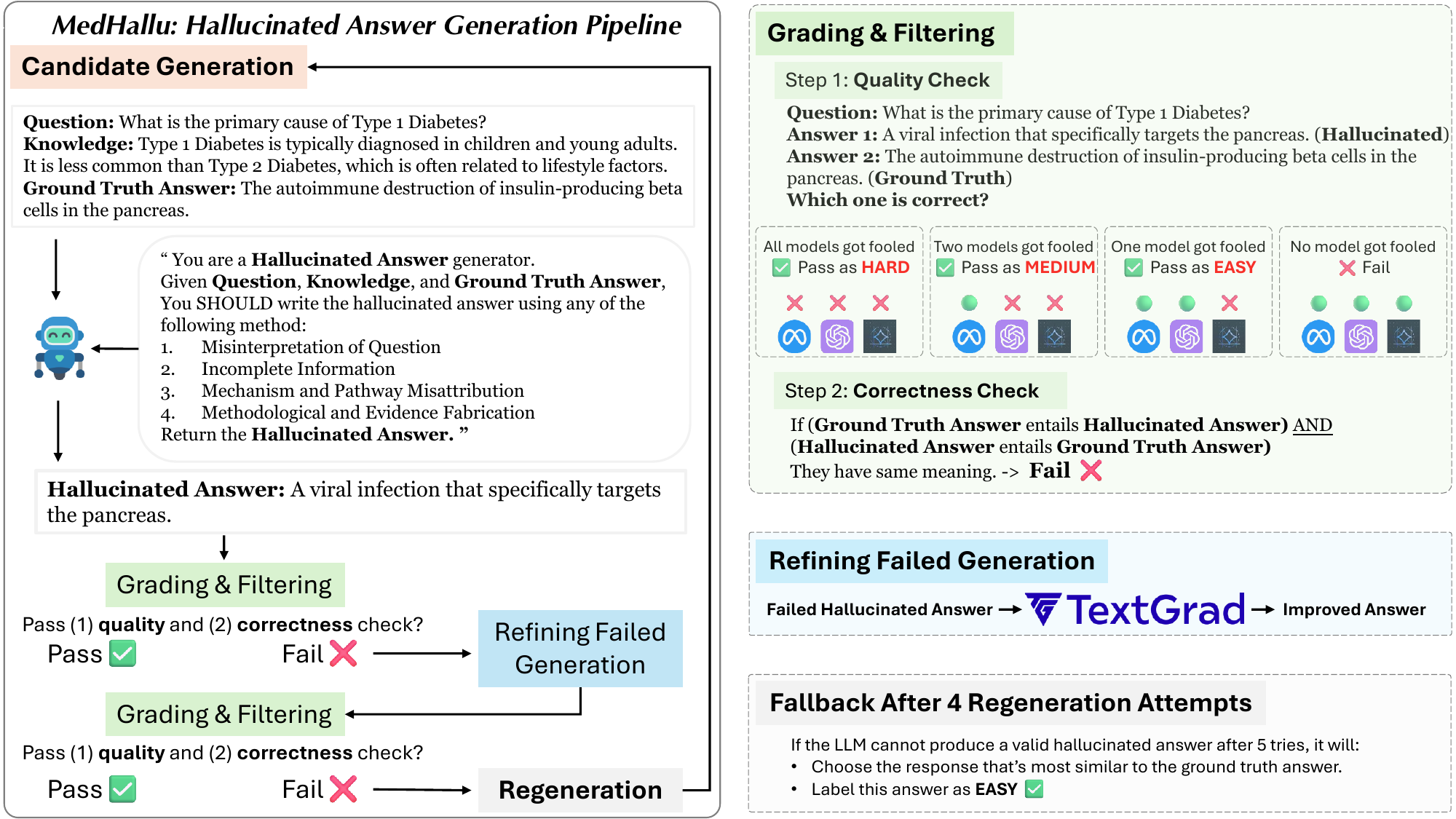}
\caption{\textbf{MedHallu} medical hallucinated answer generation pipeline. Each question-answer pair from the PubMedQA dataset undergoes the following steps to generate a hallucinated answer: (1) \textbf{Candidate Generation}: Given a question, relevant knowledge, and ground truth answer, the LLM is prompted to generate a hallucinated answer adhering to one of four hallucination types. (2) \textbf{Grading \& Filtering}: Generated answers undergo \textbf{quality} and \textbf{correctness} checks, being labeled as \textbf{hard}, \textbf{medium}, \textbf{easy}, or \textbf{failed} based on filtering results. (3) \textbf{Refining Failed Generation}: Failed answers are optimized using TextGrad~\cite{yuksekgonul2024textgradautomaticdifferentiationtext} and re-filtered. If they fail again, the LLM is re-prompted to generate new answers (\textbf{Regeneration}). (4) \textbf{Fallback}: If no qualified answers emerge after four regeneration attempts, the answer most similar to the ground truth is selected as an easy hallucinated example. The detailed prompt used for hallucination generation task is presented in the Appendix~\ref{appendix:prompt}.}
\label{fig:pipeline}
\end{figure*}

\vspace{-3mm}
\section{Related Work}
\vspace{-2mm}
\paragraph{Hallucination Detection Benchmarks.}
 Hallucination in LLMs has been extensively documented in a variety of tasks, including machine translation~\citep{lee2019hallucinations_13RW}, dialogue systems~\citep{balakrishnan-etal-2019-constrained_14RW}, text summarization~\citep{durmus-etal-2020-feqa_15RW}, and question answering~\citep{sellam2020bleurtlearningrobustmetrics_16RW}, as reviewed in recent surveys~\citep{Ji_2023_12RW}. Existing benchmarks for hallucination detection, such as Hades~\citep{liu2022tokenlevelreferencefreehallucinationdetection} and HaluEval~\citep{Hallueval}, offer robust methodologies for identifying hallucinated content. However, they predominantly employ generic techniques that fail to account for the nuanced complexities inherent in medical contexts. Similarly, while benchmarks such as HaluBench~\citep{ravi2024lynxopensourcehallucination} include some medical data samples in their data set, their data generation processes are not specifically tailored for the medical domain. Although Med-HALT~\citep{pal2023medhaltmedicaldomainhallucination} focuses on medical hallucinations, it mainly serves as a performance evaluation tool rather than providing a structured dataset. In contrast, our work introduces the first comprehensive dataset for medical hallucination detection, employing controlled methods to address these domain-specific challenges.
\vspace{-2mm}
\paragraph{Semantic Analysis of Hallucinated Text.}
Hallucinated sentences often sound over-confident~\citep{miao2021preventlanguagemodeloverconfident_overconfident, chen2022improvingfaithfulnessabstractivesummarization_overconfident} and frequently contain tokens that are statistically improbable within a given context, primarily due to suboptimal decoding strategies. Fine-tuned models have sought to mitigate this issue by adjusting decoding parameters to enhance factual accuracy, thereby reducing the occurrence of rare or anomalous terms in hallucinated outputs~\citep{Huang_2025_survey11}. Despite these advancements, previous research has not systematically compared hallucinated sentences with their corresponding ground truth to assess semantic similarities. Our work fills this gap by uncovering deeper semantic relationships between hallucinated texts and their ground truth counterparts.
\vspace{-2mm}

\paragraph{Improvement Methods in Hallucination Detection.}

Recent advancements in hallucination detection have focused on integrating external knowledge to enhance model performance. Retrieval-augmented methods~\citep{lewis2021retrievalaugmentedgenerationknowledgeintensivenlp_241, li2023weboysterimprovinglarge_242} have mitigate hallucinations via grounding models in general knowledge. However, few studies have examined the impact of domain-specific knowledge on hallucination detection tasks. While HaluEval~\citep{Hallueval} evaluates knowledge-augmented detection, it lacks fine-grained, domain-relevant knowledge integration. LLMs often overestimate their competence \citep{zhang2023sirenssongaiocean_247}, which underscores the need for structured mechanisms to allow models to abstain from answering when uncertain. Prior works have leveraged reinforcement learning~\cite{xu2024rejectionimprovesreliabilitytraining_246}, conformal abstention~\citep{yadkori2024mitigatingllmhallucinationsconformal_245}, or likelihood score and entropy-based metrics~\citep{cole2023selectivelyansweringambiguousquestions_248} to guide refusal decisions. However, these methods rely on complex supervision or predefined thresholds. More straightforward approaches, such as refusing to answer out-of-domain questions~\citep{cao2024learnrefusemakinglarge_244}, offer greater practicality but lack adaptability to domain-specific tasks, particularly in complex fields like medicine. Our work addresses these limitations by (1) incorporating task-specific medical knowledge to enhance hallucination detection and (2) introducing a self-supervised “not sure” class, enabling models to autonomously abstain from answering when uncertain, without requiring elaborate supervision. This dual approach remains under-explored in medical NLP, where precision and reliability are paramount.

\vspace{-3mm}
\section{MedHallu Benchmark} \label{Methodology}
\begin{figure}[t]
    \centering
    \tiny
    \resizebox{\columnwidth}{!}{
        \begin{tikzpicture}[font=\small]
    \definecolor{easycolor}{RGB}{46,204,113}
    \definecolor{mediumcolor}{RGB}{52,152,219}
    \definecolor{hardcolor}{RGB}{231,76,60}
    
    \begin{axis}[
        width=10cm,
        height=6cm,
        ybar,
        bar width=8pt,
        ymin=0, ymax=0.65,
        ylabel={\% of samples by difficulty},
        symbolic x coords={MPM, II, MQ, MEF},
        xtick=data,
        xticklabels={
            {Mechanism\\Misattribution},
            {Incomplete\\Information},
            {Question\\Misinterpretation},
            {Evidence\\Fabrication}
        },
        x tick label style={align=center, font=\scriptsize},
        y tick label style={font=\scriptsize},
        legend style={
            at={(0.98,0.98)},
            anchor=north east,
            font=\scriptsize,
            legend columns=-1,
            transpose legend,
            row sep=0pt,
            column sep=5pt
        },
        ymajorgrids=true,
        grid style={dashed, gray!30},
        nodes near coords,
        every node near coord/.append style={
            anchor=south,
            font=\tiny,
            yshift=2pt
        },
        enlarge x limits=0.15,
        axis lines*=left  
    ]
        \addplot[fill=easycolor, bar shift=-0.5cm] coordinates {
            (MPM, 0.3295) (II, 0.2635) (MQ, 0.3276) (MEF, 0.3265)
        };
        \addplot[fill=mediumcolor, bar shift= 0.00cm] coordinates {
            (MPM, 0.3322) (II, 0.3290) (MQ, 0.3182) (MEF, 0.3061)
        };
        \addplot[fill=hardcolor, bar shift= 0.5cm] coordinates {
            (MPM, 0.3384) (II, 0.4075) (MQ, 0.3542) (MEF, 0.3673)
        };
        \legend{Easy, Medium, Hard}
        
        \node[anchor=south] at (axis cs: MPM, 0.46) {\scriptsize Total: 1129};
        \node[anchor=south] at (axis cs: II, 0.46) {\scriptsize Total: 1222};
        \node[anchor=south] at (axis cs: MQ, 0.46) {\scriptsize Total: 7600};
        \node[anchor=south] at (axis cs: MEF, 0.46) {\scriptsize Total: 49};
        
    \end{axis}
\end{tikzpicture}
    }
    \caption{Statistics of the MedHallu dataset categorized by four categories of hallucinations (see Table \ref{tab:medical_hallucination_types} for detailed definitions) and levels of difficulty (easy, medium, hard).}
    \label{fig:statistics}
\end{figure}
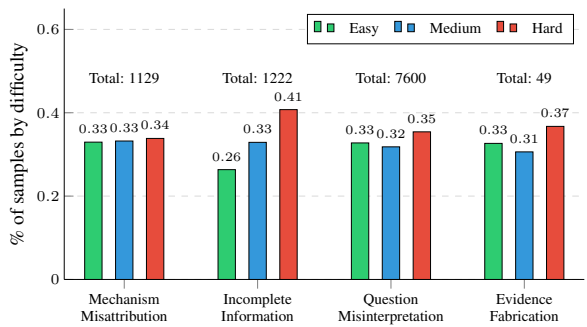

We create this dataset by proposing a simple yet effective pipeline with minimal human intervention, making it easy to scale the data generation. Figure~\ref{fig:pipeline} describes our complete generation and filtration pipeline, while Algorithm \ref{alg:one} provides a detailed approach for the same. We draw inspiration from the definitions of hallucinated answers provided by the KnowHalu paper~\citep{KnowHallu}, but modify them by adding and removing certain categories to better adapt to the medical domain. By defining the medical domain-specific hallucination categories, as presented in Table~\ref{tab:medical_hallucination_types}, we ensure that the generated dataset reflects potential hallucination in the medical domains. We present the distribution of samples by hallucination categories and levels of difficulty (Figure~\ref{fig:statistics}) for the MedHallu dataset, which consists of 10,000 samples in total. The difficulty distribution of hallucinated answers is relatively even, with the ``hard'' type being slightly more common than the ``easy'' and ``medium'' types. The distribution of hallucination categories by definition is more concentrated. Misinterpretation of the question is the most common hallucination category in MedHallu, accounting for 76\% of the entire dataset, while evidence fabrication represents the smallest portion (0.5\%).

\begin{table*}
\centering
\small
\renewcommand{\arraystretch}{1.5} 
\vspace{-2mm}
\resizebox{\textwidth}{!}{
\begin{tabular}{>{\centering\arraybackslash}m{0.18\textwidth}|>{\arraybackslash}m{0.30\textwidth}|>{\arraybackslash}m{0.45\textwidth}}
\toprule
\textbf{Hallucination Category} & \multicolumn{1}{c|}{\textbf{Description}} & \multicolumn{1}{c}{\textbf{Example}} \\ \midrule
Misinterpretation of Question & Misunderstanding the question, leading to an irrelevant response. & \textbf{\#Question\#}: Does high-dose vitamin C therapy improve survival rates in patients with sepsis? \newline \textbf{\#Answer\#}: Vitamin C is water-soluble vitamin that plays a role in immune function and collagen synthesis. \\ \hline
Incomplete Information & Stays on-topic but omits the essential details needed to fully answer the question. & \textbf{\#Question\#}: How does penicillin treat strep throat? \newline \textbf{\#Answer\#}: Penicillin kills bacteria.\\ \hline
Mechanism and Pathway Misattribution & False attribution of biological mechanisms, molecular pathways, or disease processes that contradicts established medical knowledge. & \textbf{\#Question\#}: What is the primary mechanism of action of aspirin in reducing inflammation? \newline \textbf{\#Answer\#}: Aspirin primarily reduces inflammation by blocking calcium channels in immune cells, which prevents the release of histamine and directly suppresses T-cell activation. \\ \hline
Methodological and Evidence Fabrication & Inventing false research methods, statistical data, or specific clinical outcomes. & \textbf{\#Question\#}: What is the success rate of ACL reconstruction surgery? \newline \textbf{\#Answer\#}: Recent clinical trials using quantum-guided surgical technique showed 99.7\% success rate across 10,543 patients with zero complications when using gold-infused synthetic grafts. \\
\bottomrule
\end{tabular}}
\caption{Categories of medical hallucinations used to generate the MedHalu dataset. Adapted from the KnowHallu benchmark \citep{KnowHallu} with revised categories tailored to the medical domain (Appendix \ref{sec:hallu_categories}).}
\vspace{-3mm}
\label{tab:medical_hallucination_types}
\vspace{-1mm}
\end{table*}
\vspace{-2mm}

\subsection*{Dataset Generation Pipeline}
\vspace{-2mm}
The proposed methodological framework comprises a three-phase pipeline architected for robust hallucinated sample generation (Figure~\ref{fig:pipeline}). 
The pipeline follows a sequential approach: (1) stochastic sampling of potential hallucinated responses based on in-context examples and precise definitions, (2) LLM-based quality filtering mechanisms, (3) correctness checking using bidirectional entailment and LLM prompting. (4) Sequential Improvement via TextGrad. Finally, inspired by~\citep{Hallueval}, we select the most similar sample generated, using semantic similarity in cases where a high-quality sample is not identified. This approach enables comprehensive identification and evaluation of linguistic hallucinations while minimizing false positives through multi-layered verification protocols.

\begin{algorithm}[t]
\small
\caption{Hallucination Generation Pipeline} \label{alg:one}
\textbf{Input:} Question $Q$, Knowledge Context $K$, Ground truth $G$,
Number of attempts $N$,
Generator model $M_{gen}$, Discriminator models $\{D_1, D_2, ..., D_k\}$,
TextGrad (TG) model $M_{tg}$, $fooled$ checks both quality and correctness\\
\textbf{Output:} Best hallucinated response $H^*$
\vspace{1mm}
\hrule
\vspace{1mm}
\textbf{Initialize:}\\
$\mathcal{H} \leftarrow \{\}$ \Comment{Initialize candidate set}\\
$success \leftarrow False$

\textbf{Phase 1:} Generation and Evaluation\\
\For{$i \leftarrow 1$ \textbf{to} $N$}{
    $H_i \leftarrow M_{gen}(Q, K)$ \Comment{Generate initial answer}\\
    
    \For{$j \leftarrow 1$ \textbf{to} $k$}{
        $fooled_j \leftarrow D_j(Q, H_i, G)$ \Comment{Check discriminator j}\\
        \If{$fooled_j = True$}{
            $H^* \leftarrow H_i$\\
            $success \leftarrow True$\\
            \textbf{break Phase 1}
        }
    }
    
    \If{$\neg success$}{
        $H_{improved} \leftarrow M_{tg}(H_i, Q, K)$ \Comment{TG improvement}\\
        $H_i^{\prime} \leftarrow M_{gen}(Q, K, H_{improved})$\\
        
        \For{$j \leftarrow 1$ \textbf{to} $k$}{
            $fooled_j \leftarrow D_j(Q, H_i^{\prime}, G)$\\
            \If{$fooled_j = True$}{
                $H^* \leftarrow H_i^{\prime}$\\
                $success \leftarrow True$\\
                \textbf{break Phase 1}
            }
        }
    }
    
    $\mathcal{H} \leftarrow \mathcal{H} \cup \{H_i, H_i^{\prime}\}$ \Comment{Store both attempts}
}
\textbf{Phase 2:} Fallback Selection\\
\If{$\neg success$}{
    $H^* \leftarrow \arg\max_{H \in \mathcal{H}} (\text{CosineSimilarity}(H, G))$
}
\Return{$H^*$}
\end{algorithm}

\paragraph{1) Diverse Hallucinated Answer Sampling.} \label{Sampling step}

Using a carefully crafted prompting strategy shown in Figure~\ref{fig:pipeline}, we generate multiple possible hallucinated answers with diverse temperature settings, we describe the prompt in Table~\ref{fig:system_prompt}. Through experiments, we find that allowing the model to choose the category of hallucination to apply to a given medical question performs better than manually forcing a specific hallucination category. For this generation $H_i = LM_i(Q_i,GT_i,C_i)$, we provide the LLM with precise definitions of each category, along with examples, question $Q_i$, and ground truth answers $GT_i$. The LLM is tasked with generating an answer that is semantically similar to ground truth yet incorrect. Additionally, we provide the ground truth context $C_i$, which contains precise knowledge required to answer the question. This includes intricate details necessary for crafting a strong hallucinated answer. 

\paragraph{2) Quality checking - LLM-based Discriminative Filtering.} \label{Quality check}

The second phase of our pipeline implements a comprehensive quality filtering protocol leveraging an ensemble of LLMs to minimize individual model biases. For each generated sample $H_i$, we employ a comparative assessment framework where multiple LLMs independently evaluate two candidate responses: the potentially hallucinated answer and the established ground truth. The quality assessment task is formulated as a binary classification problem, where models are prompted to identify which response appears more factually accurate given the question without access to the ground truth context. To mitigate potential biases from any single model, we implement a majority voting mechanism across different LLM architectures (including Gemma2, GPT-4o-mini, and Qwen2.5). A generated sample $H_i$ is preserved only when at least a majority of models in the ensemble incorrectly identify it as the more accurate response compared to the ground truth.
The difficulty categorization of generated samples is determined by the voting patterns across the LLM ensemble. Specifically, we classify $H_i$ as ``hard'' when all LLMs in the ensemble incorrectly identify it as accurate response, ``medium'' when multiple but not all LLMs are deceived, and ``easy'' when only a single LLM fails to identify the hallucination. This multi-model consensus approach helps ensure that preserved hallucinated samples are sufficiently convincing while reducing the impact of model-specific quirks or biases in the filtering process.
\vspace{-1mm}
\paragraph*{3) Correctness Checking via Entailment.} \label{Correctness check}
We implement a two-stage correctness verification protocol to ensure that the generated hallucinations are semantically distinct from the ground truth while maintaining coherence. First, we employ bidirectional entailment checking using a fine-tuned RoBERTa-large-MNLI model to quantify the semantic divergence between the hallucinated sample $H_i$ and ground truth $GT_i$. The bidirectional entailment score $\mathcal{E}$ is computed as:
$$\mathcal{E}(H_i, GT_i) = \min(\text{NLI}(H_i \rightarrow GT_i), \text{NLI}(GT_i \rightarrow H_i))$$

where $\text{NLI}(x \rightarrow y)$ represents the natural language inference score indicating whether $x$ entails $y$. We establish a stringent threshold $\tau$ and only retain samples that satisfy: $\mathcal{E}(H_i, GT_i) < \tau$. This ensures the hallucinated samples maintain sufficient semantic distance from the ground truth, minimizing false positives while requiring minimal human intervention.

\paragraph*{4) Sequential Improvement via TextGrad.} \label{improvement}
Our framework implements an iterative optimization step to enhance the quality of generated hallucinations that fail initial quality or correctness checks. When a generated sample $H_i$ fails to meet the established quality tests described in Section~\ref{Quality check}, we employ TextGrad optimization to refine subsequent generations through a feedback loop. The optimization process is formalized as: $H_{i+1} = \text{TextGrad}(H_i, F(H_i))$ where $F(H_i)$ represents feedback from the TextGrad optimizer, initialized with GPT-4o-mini. This refinement process (detailed in Section~\ref{Sampling step}) iterates up to five times, terminating either upon reaching a quality-compliant sample or exhausting the iteration limit. For each failed generation, TextGrad analyzes LLM feedback to identify hallucination indicators that make $H_i$ easily detectable. The feedback mechanism specifically focuses on two aspects: (1) linguistic patterns that signal artificial content and (2) structural elements that could be refined to enhance the naturalness. This feedback is then incorporated into subsequent prompt refinement, systematically improving both the content plausibility and stylistic cohesion. If no sample passes the quality filter after maximum iterations, we implement a fallback strategy based on semantic dissimilarity. Specifically, we select the candidate $H_*$ that maximizes the cosine similarity from the ground truth embedding: $H_* = \arg\max_{H_i} (\cos(\text{embed}(H_i), \text{embed}(GT_i)))$. This ensures that even in challenging cases, our pipeline produces outputs with maximum semantic similarity while preserving response coherence.

\begin{table*}[t]
    \centering
    \renewcommand{\arraystretch}{1.2}
    \begin{adjustbox}{width=1\linewidth, center}
    \begin{tabular}{lccccc|ccccc|c}
    \toprule
    \textbf{Model} & \multicolumn{5}{c|}{\textbf{Without Knowledge}} & \multicolumn{5}{c|}{\textbf{With Knowledge}} & \textbf{$\Delta$ Knowledge} \\
    \cmidrule(lr){1-6} \cmidrule(lr){7-11}
    \textbf{General LLMs} & \textbf{Overall F1} & \textbf{Overall P} & \textbf{Easy F1} & \textbf{Med F1} & \textbf{Hard F1} 
                   & \textbf{Overall F1} & \textbf{Overall P} & \textbf{Easy F1} & \textbf{Med F1} & \textbf{Hard F1} & ($\Delta$ F1)\\
    \cmidrule(lr){2-6} \cmidrule(lr){7-12}
    GPT-4o$^*$                       & \textbf{0.737} & 0.723 & \textbf{0.844} & \textbf{0.758} & \textbf{0.625} & \textbf{0.877} & \textbf{0.882} & \textbf{0.947} & \textbf{0.880} & \textbf{0.811} & 0.140 \\
    GPT-4o mini                  & 0.607 & 0.772 & 0.783 & 0.603 & 0.446 & 0.841 & 0.820 & 0.914 & 0.854 & 0.761 & 0.234 \\
    Qwen2.5-14B-Instruct         & 0.619 & 0.691 & 0.773 & 0.611 & 0.483 & 0.852 & 0.857 & 0.935 & 0.856 & 0.769 & 0.233 \\
    Gemma-2-9b-Instruct          & 0.515 & 0.740 & 0.693 & 0.512 & 0.347 & 0.838 & 0.809 & 0.918 & 0.848 & 0.758 & \textbf{0.323} \\
    Llama-3.1-8B-Instruct        & 0.522 & \textbf{0.791} & 0.679 & 0.515 & 0.372 & 0.797 & 0.775 & 0.880 & 0.796 & 0.722 & 0.275 \\
    DeepSeek-R1-Distill-Llama-8B & 0.514 & 0.570 & 0.589 & 0.515 & 0.444 & 0.812 & 0.864 & 0.895 & 0.794 & 0.751 & 0.298 \\
    Qwen2.5-7B-Instruct          & 0.553 & 0.745 & 0.733 & 0.528 & 0.402 & 0.839 & 0.866 & 0.923 & 0.832 & 0.770 & 0.286 \\
    Qwen2.5-3B-Instruct          & 0.606 & 0.495 & 0.667 & 0.602 & 0.556 &  0.676 & 0.514 & 0.693 & 0.677 & 0.661 & 0.070 \\
    Llama-3.2-3B-Instruct                  & 0.499 & 0.696 & 0.651 & 0.467 & 0.384 & 0.734 & 0.775 & 0.822 & 0.723 & 0.664 & 0.235 \\
    Gemma-2-2b-Instruct                   & 0.553 & 0.620 & 0.680 & 0.524 & 0.457 & 0.715 & 0.786 & 0.812 & 0.705 & 0.631 & 0.162 \\
    
    \midrule
    
    \textbf{Medical Fine-Tuned LLMs} & \textbf{Overall F1} & \textbf{Overall P} & \textbf{Easy F1} & \textbf{Med F1} & \textbf{Hard F1} 
                   & \textbf{Overall F1} & \textbf{Overall P} & \textbf{Easy F1} & \textbf{Med F1} & \textbf{Hard F1} & {($\Delta$ F1)}\\
    \cmidrule(lr){2-6} \cmidrule(lr){7-12}
    OpenBioLLM-Llama3-8B         & 0.484 & 0.490 & 0.494 & 0.474 & 0.483 & 0.424 & 0.567 & 0.438 & 0.412 & 0.423 & -0.060 \\
    BioMistral-7B                & 0.570 & 0.518 & 0.627 & 0.563 & \textbf{0.525} & 0.648 & 0.516 & 0.652 & 0.660 & 0.634 & 0.078 \\
    Llama-3.1-8B-UltraMedical    & \textbf{0.619} & 0.657 & \textbf{0.747} & \textbf{0.596} & 0.524 & 0.773 & 0.679 & 0.832 & 0.777 & \textbf{0.718} & 0.153 \\
    Llama3-Med42-8B & 0.416 & \textbf{0.829} & 0.600 & 0.379 & 0.264 & \textbf{0.797} & \textbf{0.856} & \textbf{0.898} & \textbf{0.794} & 0.707 & \textbf{0.381} \\
    \midrule
    \textbf{Average (General LLMs, w/o GPT-4o)} 
                                 & \textbf{0.533} & \textbf{0.686} & \textbf{0.674} & \textbf{0.517} & 0.412 
                                 & \textbf{0.784} & \textbf{0.789} & \textbf{0.864} & \textbf{0.781} & \textbf{0.716} 
                                 & \textbf{0.251} \\
    \textbf{Average (Medical Fine-Tuned LLMs)} 
                                 & 0.522 & 0.623 & 0.617 & 0.503 & \textbf{0.449} 
                                 & 0.660 & 0.654 & 0.705 & 0.660 & 0.620 
                                 & 0.138 \\
    \bottomrule
    \end{tabular}
    \end{adjustbox}
    \caption{Performance comparison of different LLMs with and without knowledge on MedHallu (10,000 samples). General LLMs perform better than medically fine-tuned LLMs in the task of Medical Hallucination across most metrics. “Overall P” denotes precision, and “\(\Delta\) Knowledge” is the performance change in overall F1 when knowledge is provided. $^*$We exclude GPT-4o while calculating the average to have a fair comparison of model sizes for general vs. fine-tuned LLMs. Additional experimental details can be found in Appendix \ref{Gen_robust_check}.}
    \label{tab:llm_performance_merged}
\end{table*}
\vspace{-2.5mm}
\section{Implementation Details}
\vspace{-2.5mm}
\paragraph{MedHallu Dataset Generation Settings.} We generate hallucinated responses using \texttt{Qwen2.5B-14B}~\citep{qwen2025qwen25technicalreport}. The ground truth question-answer pairs are sourced from the \texttt{pqa\_labeled} split of PubMedQA~\citep{PubmedQA}, which contains 1,000 expert-annotated samples, supplemented with 9,000 instances randomly selected from the \texttt{pqa\_artificial} split. To achieve high-quality generation with adequate diversity, we utilize regulated sampling settings. The \texttt{temperature} is varied between 0.3 and 0.7, while the \texttt{nucleus sampling threshold (top-p)} is fixed at 0.95. These settings balance cohesion and variability. The maximum response length is capped at 512 tokens to ensure completeness while mitigating computational costs. Each hallucinated answer is limited to within ±10\% of its corresponding ground truth answer's length, ensuring uniform information density.

\noindent $\triangleright$ \textit{Quality \& correctness check.} For quality check, We employ three LLMs: \texttt{GPT-4o mini}~\citep{openai2024gpt4ocard}, \texttt{Gemma2-9B}~\citep{gemmateam2024gemma2improvingopen}, and \texttt{Qwen2.5-7B}~\citep{qwen2025qwen25technicalreport}. A response is retained only if it deceives at least one of these models (see Section~\ref{Quality check}). For \textit{correctness check,} we employ the \texttt{microsoft/deberta-large-mnli} model~\citep{he2021debertadecodingenhancedbertdisentangled},  applying bidirectional entailment with a confidence threshold of 0.75. 

\noindent $\triangleright$ \textit{TextGrad \& Fallback.} We integrate \texttt{TextGrad}~\cite{yuksekgonul2024textgradautomaticdifferentiationtext} with \texttt{GPT-4o mini} as the backend model to generate feedback for samples that fail either the quality or correctness checks. Each sample undergoes a maximum of five generation attempts. If no valid response is produced within these iterations, we adopt a \textbf{fallback strategy}, selecting the most semantically similar generated answer to the ground truth response.

\paragraph{Discriminator Model Settings.} We evaluate a diverse set of model architectures under two distinct settings: (1) \textbf{zero-shot} (without additional knowledge) and (2) \textbf{context-aware} (with ground truth context provision). The detection prompt is described in Figure~\ref{fig:system_prompt_for_detection}. This dual-setting approach allows us to assess both the baseline detection capabilities and the models' ability to leverage contextual information for improved discrimination. We examine both general-purpose and specialized medical models. The \textbf{general models} include \texttt{Gemma-2 (2B, 9B) Instruct}~\citep{gemmateam2024gemma2improvingopen}, \texttt{Llama-3.1 (3B, 8B) Instruct}~\citep{grattafiori2024llama3herdmodels}, \texttt{Qwen-2.5 (3B, 7B, 14B)}~\citep{qwen2025qwen25technicalreport}, \texttt{DeepSeek-R1-Llama 8B}~\citep{deepseekai2025deepseekr1incentivizingreasoningcapability}, \texttt{GPT-4o}, and \texttt{GPT-4o mini}~\citep{openai2024gpt4ocard}. Additionally, we evaluate four \textbf{fine-tuned medical LLMs} such as \texttt{OpenBioLLM-8B}~\citep{OpenBioLLMs}, \texttt{Llama3-Med42-8B}~\citep{christophe2024med42v2suiteclinicalllms}, \texttt{BioMistral-7B}~\citep{labrak2024biomistral}, and \texttt{UltraMedical}~\citep{zhang2024ultramedical} to compare domain-specific adaptations against general-purpose models. In this discriminative task, we maintain a temperature of approximately \texttt{0.2-0.3} for all models. For OpenAI models, we use the official API, while for open-weight models like Llama, Gemma, and Qwen, we utilize the \texttt{Hugging Face Pipeline} to ensure a consistent inference framework across all models.


\section{Results and Analysis} \label{ResultsAnalysis}



\vspace{-2mm}
\subsection{Which language model performs the best at medical hallucination detection task?}

Our experimental results reveal significant variations in hallucination detection performance across model architectures in the zero-shot setting (without relevant knowledge provided). As presented in Table~\ref{tab:llm_performance_merged}, \ding{202}~the size of a model is not necessarily linked to its detection capabilities. For instance, \texttt{Qwen2.5-3B} achieves a high baseline overall F1 score (0.606), outperforming larger models such as \texttt{Gemma-9B} (0.515), \texttt{Llama-3.1-8B-Instruct} (0.522), and even the \texttt{Qwen2.5-7B} model (0.533). \ding{203}~All models exhibit notable performance degradation on ``hard'' samples, with even the best-performing models, such as \texttt{GPT-4o}, showing a significant F1 score drop and achieving only 0.625 in these challenging cases. \ding{204}~An intriguing observation is that, overall, general LLMs outperform medical fine-tuned LLMs in terms of precision and F1 scores in the easy and medium categories when no additional knowledge is provided.

\vspace{-1mm}
\subsection{How does providing knowledge impact detection performance?}
Providing knowledge to the LLMs in this hallucination detection task, yields substantial and consistent improvements in hallucination detection across all evaluated LLM architectures. As shown in Table \ref{tab:llm_performance_merged}, \ding{202} every model benefits from the inclusion of knowledge. In general LLMs, the average overall F1 score increases from 0.533 (without knowledge) to 0.784 (with knowledge), corresponding to a gain of +0.251. In contrast, medically fine-tuned LLMs exhibit a much smaller improvement—from an average overall F1 of 0.522 to 0.660 (+0.138), likely because these models already incorporate specialized domain knowledge during training. Moreover, the scale of the model is pivotal for its performance. \ding{203} Larger structures, such as \texttt{Qwen2.5-14B}, reach an impressive overall F1 score of 0.852 when supplemented with domain knowledge, indicating that their increased capacity supports better text comprehension and integration of knowledge. In contrast, smaller models like \texttt{Qwen2.5-3B} experience just slight enhancement (+0.07 F1, from 0.606 to 0.676), underscoring the variability in how different model sizes effectively use additional information. Remarkably, \texttt{Gemma-2-9B} showed the most significant benefit from knowledge, with its overall F1 score rising from 0.515 to 0.838 (+0.323). Overall, these findings affirm the hypothesis that domain knowledge access improves an LLM's hallucination detection ability, while also emphasizing that both model scale and whether the model has been fine-tuned on medical data are critical to the extent of performance improvements.

\begin{table}
\centering
\small
\begin{tabular}{lccc}
\hline
\textbf{Metric} & \textbf{Mean} & \textbf{Mean} & \textbf{P-value} \\
 & \textbf{(fooled)} & \textbf{(not fooled)} & \\
\hline
Cosine similarity & 0.715 & 0.696 & 0.004 \\
Euclidean distance & 0.714 & 0.750 & 0.002 \\
Rouge1-F1 & 0.358 & 0.319 & 0.002 \\
\hline
\end{tabular}
\caption{The average similarity between the clusters generated in Section \ref{semantic_analysis} and the ground truth samples. Clusters containing samples that fool detection LLMs (i.e., hallucinations that are more challenging to detect) are notably closer to the ground truth.}
\label{tab:metrics_cluster}
\end{table}

\subsection{Semantic analysis of hallucinated and ground truth sentences.} \label{semantic_analysis}
\vspace{-1mm}
To analyze semantic patterns in hallucinated responses, we conduct a comprehensive clustering analysis on an expanded set of generations. Specifically, we generate 50 candidate hallucinated responses for each question from our sampling phase, as described in Section~\ref{Quality check}. We retain all 50 candidate hallucinated responses, including those that fail the quality or correctness checks, to capture the semantic distribution across both successful and unsuccessful hallucinated answers. Using bidirectional entailment with a threshold of 0.75, we cluster these 50 candidate hallucinated responses along with the ground truth response, forming distinct semantic clusters that represent different conceptual approaches to the same question. This clustering methodology, adapted from~\citep{Farquhar2024DetectingHI_semantic_entropy}, allows us to analyze the semantic structure of hallucinated responses relative to the ground truth, yielding three significant findings:

\paragraph{Cluster-level Detection Patterns.}
Our analysis uncovers a binary discrimination effect within semantic clusters. \ding{202} Specifically, hallucinated responses in the same cluster tend to exhibit near-uniform performance—either consistently passing LLM detection (being favored over the ground truth) or being uniformly flagged as hallucinations. This finding strongly indicates that semantic content, rather than merely surface-level linguistic features, plays a pivotal role in shaping the LLM's discrimination behavior.
\vspace{-1mm}
\paragraph{Cluster Proximity Analysis.}
\ding{203} We find that clusters containing samples that reliably fool detection LLMs (hallucinations that are harder to detect) are notably closer to the ground truth answer in semantic vector space. This closeness is quantified via Euclidean distance, with additional support from cosine similarity and ROUGE scores (Table~\ref{tab:metrics_cluster}). Such proximity suggests that well-crafted hallucinated responses strike a balance, they remain semantically similar enough to the ground truth while incorporating meaningful deviations.

\vspace{-1mm}
\paragraph{Ground Truth Isolation.}
A particularly significant finding is the distinct semantic isolation of ground truth responses from clusters of hallucinated outputs. Empirical evidence demonstrates that ground truth responses rarely, if ever, align within the semantic clusters formed by hallucinations. This clear separation validates the robustness of our generation pipeline, ensuring that hallucinated responses retain semantic distinctness from factual content while upholding contextual relevance.
\begin{table}[h]
\centering
\scriptsize
\begin{tabular}{@{}l c c c c c@{}}
\toprule
\textbf{Model} & \textbf{F1\textsubscript{NS}} & \textbf{P\textsubscript{NS}} & \textbf{F1\textsubscript{R}} & \textbf{P\textsubscript{R}} & \textbf{Response\%} \\ 
\midrule
GPT-4o-mini                     & 66.6  & 66.8 & 60.7 & 77.2 & 98.4   \\
Gemma-2-2b-it                  & 57.1  & 59.9 & 55.3 & 54.1 & 82.7   \\
Llama-3.2-3B-Instruct           & 58.1  & 68.7 & 49.9 & 63.3 & 85.9   \\
Qwen2.5-3B-Instruct             & 65.2  & 67.2 & 60.6 & 50.2 & 65.7   \\
BioMistral-7B                 & 56.5  & 50.5 & 57.0 & 51.3 & 99.2   \\
Qwen2.5-7B-Instruct             & 69.3  & \textbf{94.6} & 55.3 & 73.7 & 47.5   \\
OpenBioLLM-Llama3-8B            & 48.8  & 48.4 & 48.4 & 56.3 & 99.7   \\
Llama-3.1-8B-UltraMedical       & 58.5  & 49.1 & 61.9 & 56.4 & 69.7   \\
DeepSeek-R1-Llama-8B    & 66.0  & 56.9 & 51.4 & 61.7 & 98.1   \\
Llama-3.1-8B-Instruct           & 51.7  & 90.4 & 52.2 & \textbf{86.0} & 98.2   \\
Gemma-2-9b-it                  & 61.4  & 85.5 & 51.5 & 71.5 & 37.6   \\
Qwen2.5-14B-Instruct            & 76.2  & 82.9 & 61.9 & 76.5 & 27.9   \\
GPT-4o                         & \textbf{79.5}  & 79.6 & \textbf{73.7} & 72.4 & 33.9   \\
\bottomrule
\end{tabular}
\caption{F1\textsubscript{NS} and P\textsubscript{NS} (Precision) represent performance with the ``Not Sure'' option, while F1\textsubscript{R} and P\textsubscript{R} (Precision) represent performance when required to answer. Response\% represents the percentage of questions answered with ``Yes'' or ``No'' even when the ``Not Sure'' option is available.}
\label{tab:model-comparison}
\end{table}

\vspace{-5mm}
\subsection{Analysis of models' ability to decline to answer}

We introduce a ``not sure'' category alongside the existing ``hallucinated'' and ``not hallucinated'' categories in our detection prompt (Figure \ref{fig:system_prompt_for_detection}), allowing LLMs to decline to answer if they lack full confidence in their responses. Results shown in Table \ref{tab:model-comparison}, reveal that \ding{202} many models demonstrate an improved F1 score and precision when they can opt for ``not sure.'' However, the enhancement varies with model size: smaller models gain a moderate improvement of 3-5\%, whereas larger models see a significant boost of around 10-15\%. General LLMs outperform fine-tuned medical models, with some like GPT-4o achieving up to 79.5\% in performance, and Qwen2.5-14B performing closely at 76.2\%. \ding{203} In terms of the percentage of questions answered with definite ``yes'' or ``no'' (Response Rate), general LLMs respond to fewer questions, with Qwen2.5-14B responding to as little as 27.9\%, reflecting their tendency to skip uncertain questions. Conversely, fine-tuned medical models attempt to answer nearly all questions, rarely selecting the ``Not Sure'' option. This approach sometimes leads to a minor reduction in performance. For instance, UltraMedical's model has the lowest response rate among medical models at 69.7\% , while OpenBioLLM reaches as high as 99.7\%. Finally, \ding{204} when comparing the impact of adding the ``not sure'' choice with knowledge-sharing enhancements, shown in Table \ref{tab:model-comparison-with-knowledge} versus Table \ref{tab:model-comparison}, there is a marked increase in the percentage of questions attempted by General LLMs, suggesting improved confidence in task execution, along with an increase in F1 score and precision.

\begin{figure}
    \centering
    \tiny
    \resizebox{\columnwidth}{!}{
        \begin{tikzpicture}[font=\small]
    \definecolor{easycolor}{RGB}{46,204,113}
    \definecolor{mediumcolor}{RGB}{52,152,219}
    \definecolor{hardcolor}{RGB}{231,76,60}
    \definecolor{correctcolor}{RGB}{46,204,113}
    \definecolor{incorrectcolor}{RGB}{231,76,60}
    
    \begin{axis}[
        at={(0,-2cm)},
        anchor=north west,
        width=10cm,
        height=6cm,
        ybar,
        bar width=10pt,
        ymin=0, ymax=1,
        ylabel={Category-wise accuracy (\%)},
        symbolic x coords={MPM, II, MQ, MEF},
        xtick=data,
        xticklabels={
            {Mechanism\\Misattribution\\(MPM)},
            {Incomplete\\Information\\(II)},
            {Question\\Misinterpretation\\(MQ)},
            {Evidence\\Fabrication\\(MEF)}
        },
        x tick label style={align=center, font=\scriptsize},
        y tick label style={font=\scriptsize},
        legend style={
            at={(0.8,0.98)}, 
            anchor=north, 
            font=\scriptsize, 
            legend columns=-1,
            transpose legend,
            row sep=0pt,
            column sep=5pt
        },
        title={Performance of different hallucination types},
        title style={font=\small},
        ymajorgrids=true,
        grid style={dashed, gray!30},
        nodes near coords,
        every node near coord/.append style={
            anchor=south,
            font=\tiny,
            yshift=2pt
        },
        enlarge x limits=0.15,
        axis lines*=left  
    ]
        \addplot[fill=correctcolor] coordinates {
            (MPM, 0.6953) (II, 0.5376) (MQ, 0.6361) (MEF, 0.6939)
        };
        \addplot[fill=incorrectcolor, bar shift=0.3cm] coordinates {
            (MPM, 0.3047) (II, 0.4624) (MQ, 0.3639) (MEF, 0.3061)
        };
        \legend{Correct, Incorrect}
    \end{axis}
\end{tikzpicture}
    }
    \caption{Detection accuracy of different hallucination categories on MedHallu, evaluated using \texttt{Qwen2-7B-Instruct} as the discriminator.}
    \label{fig:hallucination_types}
\end{figure}
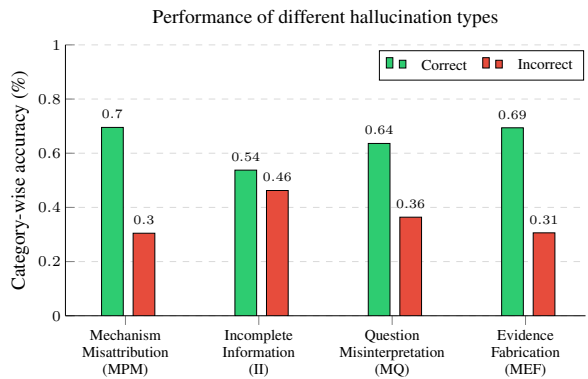

\subsection{Analysis: Hallucination category and MeSH}

\subsubsection*{Which hallucination category is hardest to detect? }
Our analysis reveals distinct patterns in detection difficulty across hallucination categories, as shown in Figure~\ref{fig:hallucination_types}. \textbf{Incomplete Information (II)} emerges as the most challenging category, with 41\% of total samples being ``hard'' cases (Figure \ref{fig:statistics}) and the lowest detection ratio (54\%), indicating models struggle significantly with validating partial information. \textbf{Mechanism/Pathway Misattribution (MPM)} and \textbf{Question Misinterpretation (MQ)} show notable patterns: MPM has a significant number of hard cases, with a 68\% detection accuracy, while MQ having higher number of hard cases but stronger detection performance (68.8\%). \textbf{Methodological and Evidence Fabrication (MEF)}, despite being the smallest category (37\% are hard), demonstrates the highest detection success rate (76.6\%).

\noindent These findings highlight a crucial insight: subtle manipulation of existing medical information, particularly through incomplete presentation, is harder to detect than outright fabrication. This is evident from II's high difficulty scores compared to MEF's better detection rates. The distribution across difficulty levels (easy, medium, hard) further supports this, with II showing the highest concentration in the ``hard'' category. This suggests that while models excel at identifying completely fabricated information, they struggle with partially accurate yet incomplete medical claims, highlighting critical areas of improvement in hallucination detection systems.

\subsubsection*{Which medical category (MeSH term) hallucination is the hardest to detect?}

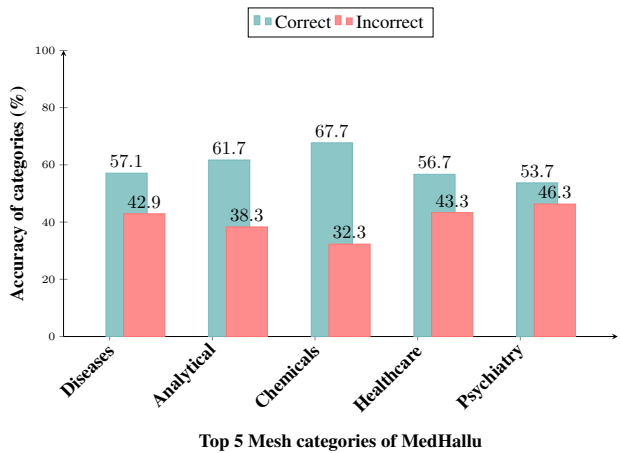
\begin{figure}[t]
    \centering
    \tiny
    \scalebox{0.7}{
        \begin{tikzpicture}
\tiny
\begin{axis}[
    width=12cm,
    height=7cm,
    ybar=-40pt,    
    bar width=22pt,
    ylabel={\normalsize\textbf{Accuracy of categories (\%)}},
    xlabel={\normalsize\textbf{Top 5 Mesh categories of MedHallu}},
    xlabel style={yshift=-1em}, 
    symbolic x coords={D1,D2,A1,A2,C1,C2,H1,H2,P1,P2},
    xtick={D1,A1,C1,H1,P1},
    xticklabels={\normalsize\textbf{Diseases}, \normalsize\textbf{Analytical}, \normalsize\textbf{Chemicals}, \normalsize\textbf{Healthcare}, \normalsize\textbf{Psychiatry}},
    xticklabel style={
        rotate=45,           
        anchor=east,         
        yshift=-0.5em        
    },
    legend style={
        at={(0.5,1.15)},
        anchor=north,
        legend columns=2,
        font=\normalsize
    },
    ymin=0,
    ymax=100,   
    axis lines=left,
    clip=false,
    enlarge x limits=0.1,  
    nodes near coords,
    nodes near coords style={font=\normalsize},
    every node near coord/.append style={yshift=1pt}
]


\addplot[fill=teal!45!white, draw=teal!55!white] coordinates {
    (D1,57.1) (A1,61.7) (C1,67.7) (H1,56.7) (P1,53.7)
}; \addlegendentry{Correct}

\addplot[fill=red!45!white, draw=red!55!white] coordinates {
    (D2,42.9) (A2,38.3) (C2,32.3) (H2,43.3) (P2,46.3)
}; \addlegendentry{Incorrect}

\label{Mesh_class_plot}
\end{axis}
\end{tikzpicture}
    }
    \caption{Detection accuracy across Mesh categories proposed in PubMedQA. We use \texttt{Qwen2.5-7B-Instruct} as a discriminator for the 1k samples of MedHallu generated on pqa\_labeled split.}
    \label{fig:mesh_categories}
\end{figure}

\noindent To understand which medical domains are more susceptible to hallucination, we examine the MedHallu dataset with the MeSH categories within the PubMedQA dataset, identifying the top five principal categories shown in Figure~\ref{fig:mesh_categories}. These categories include Diseases (comprising 25.9\% of the samples), Analytical Procedures (20.1\%), Chemical/Drug Queries (15.8\%), Healthcare Management (9.7\%), and Psychiatric Conditions (6.7\%). Detection performance among these categories varies considerably: Disease-related instances exhibit a respectable detection accuracy of 57.1\%, despite the abundance of related medical literature in the corpus. Conversely, Chemical/Drug queries demonstrate the highest detection rate at 67.7\%. In contrast, Psychiatry ranks lowest among the top five categories with a detection rate of just 53.7\%, highlighting the need for further incorporation of this data in the training corpus.

\vspace{-2mm}
\section{Conclusion}
\vspace{-2mm}
We introduce MedHallu, a comprehensive benchmark comprising 10,000 rigorously curated medical question-answer pairs with hallucinated answers. MedHallu integrates fine-grained categorization of medical hallucination types, a hallucination generation framework that balances difficulty levels while mitigating single-LLM bias through multi-model majority voting, and systematically evaluates diverse LLM configurations' hallucination detection capabilities. Our evaluation reveals that existing LLMs exhibit significant limitations in detecting medical hallucinations, particularly struggling with "hard" hallucination answers, which are closer in distance to the ground truth. We also provide insights into enhancing LLMs' hallucination detection: when knowledge is provided, general-purpose LLMs can outperform medical fine-tuned models, and allowing models to decline to answer by providing a "not sure" option improves precision in critical applications. As the largest open medical hallucination benchmark to date, MedHallu serves as a valuable resource for evaluating LLMs' medical hallucination detection abilities and offers insights into the cautious use of LLMs in high-stakes medical domains.

\section{Limitations}
Our study faces three primary constraints. First, due to resource constraints, we could not employ the most advanced reasoning models (e.g., OpenAI o1, Gemini 2.0, DeepSeek-R1) for benchmark generation. While our pipeline incorporates multi-stage LLM quality checks and regeneration steps, using state-of-the-art models would incur prohibitive computational costs. Second, our evaluation of LLMs was restricted to input-output prompting (zero-shot, with/without knowledge provision); resource limitations precluded exploration of advanced techniques like chain-of-thought or self-consistency, which might better elicit model capabilities. Third, our hallucination generation pipeline relied on the PubMedQA corpus to ensure contextual fidelity. While this ensures biomedical relevance, future work should incorporate diverse high-quality corpora to improve scalability and domain coverage.

\section{Ethics Statement}
This research adheres to rigorous ethical standards in dataset creation and evaluation. The MedHallu benchmark utilizes publicly available PubMedQA data under MIT licenses, ensuring proper attribution and compliance with source terms of use. Patient privacy is preserved through the exclusive use of de-identified biomedical literature. While our work aims to improve AI safety in healthcare, we acknowledge potential dual-use risks and advocate for responsible deployment of medical LLMs with human oversight. The benchmark's stratification enables targeted mitigation of dangerous ``hard'' hallucinations that most closely resemble factual content. All artifacts will be released with detailed documentation to promote transparency and reproducibility in medical AI safety research.

\bibliography{bib} 

\clearpage

\appendix
\section*{Appendices}

\section{Additional Related Work}
\paragraph{General LLMs vs Fine-tuned LLMs in Hallucination Detection.}
Extensive research has investigated hallucination in texts generated by pre-trained and domain-specific fine-tuned LLMs. Studies have revealed that fine-tuned LLMs exhibit reduced hallucination in text generation compared to their general-purpose counterparts~\citep{azaria2023internalstatellmknows_detect2, xiong2024llmsexpressuncertaintyempirical, arteaga2024hallucinationdetectionllmsfast}. However, despite these advancements, there remains a notable gap that no prior work has systematically evaluated the performance of domain-specific fine-tuned LLMs on hallucination detection tasks. Lynx~\citep{ravi2024lynxopensourcehallucination}, a model specifically designed for hallucination detection, has demonstrated superior performance over general-purpose LLMs across diverse datasets. Nevertheless, this study did not extend its evaluation to include LLMs fine-tuned for specialized domains, such as medicine or finance. To address this limitation, our work conducts a comparative analysis of several fine-tuned medical LLMs in the context of medical hallucination detection.

\paragraph{Evaluation of Hallucinations and Faithfulness}
The hallucination phenomenon in LLMs manifests as the production of content that lacks proper substantiation through contextual evidence or verified knowledge bases. This can be categorized into two distinct forms: factuality hallucination, which involves deviations from established real-world facts, and faithfulness hallucination, which occurs when the model's generated content diverges from the provided input context or prompt \citep{Huang_2025_survey11}. These dual manifestations represent significant challenges in ensuring the reliability and accuracy of LLM-generated outputs. There have been recent works in detecting the faithfulness of an LLM with the use of context \citep{ming2024faithevallanguagemodelstay_fact4} or even checking the faithfulness of LLMs in the absence of context \citep{roller2020recipesbuildingopendomainchatbot_fact1,min2023factscorefinegrainedatomicevaluation_fact2,chern2023factoolfactualitydetectiongenerative_fact3, wei2024longformfactualitylargelanguage_fact5}. Contrary to faithfulness, hallucinations are detected mainly focusing on the output of the LLMs rather than the context \citep{Hallueval,liu2022tokenlevelreferencefreehallucinationdetection,hu2024refcheckerreferencebasedfinegrainedhallucination}. 

\section{Incorporating Knowledge into the Analysis of Models’ Denial Capabilities}
We evaluate the setting where the model is given a choice of answering ``not sure'' when it lacks confident to answer (Table~\ref{tab:model-comparison}). We also provide the relevant knowledge in the prompt (Appendix~\ref{appendix:prompt}). The results in Table~\ref{tab:model-comparison-with-knowledge} clearly indicate the improvement in models' capability to answer the questions compared to the previous knowledge-disabled setting. Here \texttt{Qwen2.4-14B} surpasses all other models in terms of F1 and even precision. The results indicate that even though models' performance in terms of F1 increases slightly or even remains nearly similar, the precision of these models is generally improved. 

\begin{table}[t]
\centering
\scriptsize
\begin{tabular}{@{}l c c c c c@{}}
\toprule
\textbf{Model} & \textbf{F1\textsubscript{NS}} & \textbf{P\textsubscript{NS}} & \textbf{F1\textsubscript{R}} & \textbf{P\textsubscript{R}} & \textbf{Response \%} \\ 
\midrule
GPT-4o-mini                & 83.6 & 77.7 & 84.1 & 82.0 & 100.0 \\
Gemma-2-2b-it             & 75.5 & 67.2 & 71.5 & 67.4 & 89.2  \\
Llama-3.2-3B-Instruct       & 76.8 & 67.9 & 73.4 & 55.5 & 90.8  \\
Qwen2.5-3B-Instruct         & 69.2 & 47.0 & 67.6 & 49.8 & 94.2  \\
BioMistral-7B             & 67.2 & 53.2 & 64.8 & 54.5 & 98.7  \\
Qwen2.5-7B-Instruct         & 88.6 & 91.6 & 83.9 & 85.0 & 74.6  \\
OpenBioLLM-Llama3-8B        & 40.2 & 58.9 & 42.4 & 55.5 & 99.4  \\
Llama-3.1-8B-UltraMedical   & 72.9 & 56.1 & 77.3 & 73.4 & 95.1  \\
DeepSeek-R1-Llama-8B& 68.9 & 85.4 & 81.2 & 83.4 & 95.2  \\
Llama-3.1-8B-Instruct       & 77.7 & \textbf{92.7} & 80.0 & \textbf{88.6} & 99.7  \\
Gemma-2-9b-it             & 84.7 & 83.4 & 83.8 & 78.6 & 90.3  \\
Qwen2.5-14B-Instruct        & \textbf{88.8} & 92.5 & 85.2 & 84.3 & 87.0  \\
GPT-4o                    & 84.9 & 78.3 & \textbf{87.7} & 88.3 & 95.2  \\
\bottomrule
\end{tabular}
\caption{ F1\textsubscript{NS} and P\textsubscript{NS} (Precision) represent performance with the ``Not Sure'' option, while F1\textsubscript{R} and P\textsubscript{R} represent performance when required to answer. Response\% represents the percentage of questions answered with ``Yes'' or ``No'' even when the ``Not Sure'' option is available.}
\label{tab:model-comparison-with-knowledge}
\end{table}

\section{Additional Data Correctness Check}
In addition to the existing correctness check proposed in Section~\ref{Correctness check}, we leverage Llama-3.1 to perform a lightweight semantic comparison between $H_i$ and $GT_i$. Through carefully crafted prompts, the model assesses whether the hallucinated response differs meaningfully in semantic content from the ground truth. This additional verification layer provides a cost-effective mechanism to filter out subtly similar generations that might have passed the initial entailment check.

\section{Selecting Medical Hallucination Categories for MedHallu} \label{sec:hallu_categories}
We adapted hallucination categories from KnowHallu \cite{KnowHallu} to categorize generated outputs (Table~\ref{fig:hallucination_types}). KnowHallu includes categories such as \textit{Vague}, \textit{Parroting}, and \textit{Overgeneralization}, which are more suited for hallucination detection rather than generation. These categories pose challenges in a generative setting because crafting high-quality examples that convincingly exhibit extreme parroting or subtle overgeneralization in a way that can reliably mislead a discriminator is inherently difficult. Moreover, such cases may not be as informative for evaluating generative models, as they focus on stylistic nuances rather than substantive factual inconsistencies. To ensure a more effective classification for generation, we examined various medical research papers and carefully designed a set of hallucination categories that best capture the types of errors relevant to medical text generation. This approach allows for a more meaningful evaluation of generative models while maintaining both diversity and practical relevance in the generated outputs.

\begin{table}
\centering
\begin{tabular}{p{0.9\columnwidth}}
\toprule
\textbf{Computational Budget and Infrastructure Details while generating MedHallu} \\
\midrule
\textbf{Primary Model:} Qwen2.5-14B (14B parameters) \\[1ex]
\textbf{Computation Time:} 26.5 hours \\[1ex]
\textbf{GPU Hardware:} 4 x NVIDIA RTX A6000 (48,685 MiB RAM each) \\[1ex]
\textbf{Additional Models:} Gemma2-9B (9B parameters), Qwen2.5-7B (7B parameters), GPT4omini (used for correctness check) \\[1ex]
\textbf{Dataset Size:} 10,000 samples \\
\bottomrule
\end{tabular}
\caption{Computational Budget and Infrastructure Details while generating MedHallu Dataset, not includes the discriminator models used in benchmarking.}
\label{tab:comp-details}
\end{table}

\begin{table}[t!]
\centering
\begin{tabular}{l}
\hline
\textbf{Pre-trained Models and huggingface names} \\
\hline
\texttt{m42-health/Llama3-Med42-8B} \\
\texttt{OpenMeditron/Meditron3-8B} \\
\texttt{aaditya/OpenBioLLM-Llama3-8B} \\
\texttt{BioMistral/BioMistral-7B} \\
\texttt{TsinghuaC3I/Llama-3.1-8B-UltraMedical} \\
\texttt{deepseek-ai/DeepSeek-R1-Distill-Llama-8B} \\
\texttt{Qwen/Qwen2.5-14B-Instruct} \\
\texttt{google/gemma-2-2b-it} \\
\texttt{google/gemma-2-9b-it} \\
\texttt{meta-llama/Llama-3.1-8B-Instruct} \\
\texttt{meta-llama/Llama-3.2-3B-Instruct} \\
\texttt{Qwen/Qwen2.5-7B-Instruct} \\
\texttt{Qwen/Qwen2.5-3B-Instruct} \\
\hline
\end{tabular}
\caption{List of pre-trained models with their huggingface names used in our experiments.}
\label{tab:models}
\end{table}

\section{MedHallu Creation Using Other Open-weights LLMs} \label{Gen_robust_check}

We construct the MedHallu dataset using open-weights LLMs, including \texttt{Qwen2.5-14B} and \texttt{Gemma2-9B}. Initially, we generate 1,000 samples based on the high-quality, human-annotated \texttt{pqa\_labeled\_split} from PubMedQA. To ensure quality, we employ smaller LLMs, including \texttt{GPT-4o mini}, \texttt{Gemma2-2B}, and \texttt{Llama-3.2-3B} variants, for verification. Subsequently, we evaluate various LLMs, including both general-purpose and fine-tuned medical models, on these datasets. The results for the \texttt{Gemma2-9B-IT} model are presented in Table~\ref{tab:llm_performance_gemma}, while those for the \texttt{Qwen2.5-14B} model are reported in Table~\ref{tab:llm_performance_qwen}. We conduct three independent runs for dataset generation and report the mean and standard deviation of the results. During our analysis, we observed that the Qwen model exhibited faster generation speeds and consistent generation quality with fewer cases that fail quality checks on average, thus saving up more on time and computing budget, so we decided to generate the entire dataset using \texttt{Qwen2.5-14B}. Consequently, we selected the \texttt{Qwen2.5-14B} to generate an expanded dataset comprising 10,000 samples. We see that the results in the Tables~\ref{tab:llm_performance_gemma} and \ref{tab:llm_performance_qwen} are also in alignment with the observations we presented in Section~\ref{ResultsAnalysis} of the paper, thereby bolstering our claim and contribution.

\begin{table*}[t]
    \centering
    \renewcommand{\arraystretch}{1.2}
    \begin{adjustbox}{width=1\linewidth,center}
    \begin{tabular}{lccccc|ccccc|c}
    \toprule
    \textbf{Model} & \multicolumn{5}{c|}{\textbf{Without Knowledge}} & \multicolumn{5}{c|}{\textbf{With Knowledge}} & \textbf{$\Delta$ F1} \\
    \cmidrule(lr){2-6} \cmidrule(lr){7-11}
     & Overall F1 & Overall P & Easy F1 & Med F1 & Hard F1 
       & Overall F1 & Overall P & Easy F1 & Med F1 & Hard F1 & \\
    \midrule
    \multicolumn{12}{l}{\textbf{General LLMs}} \\
    \cmidrule(lr){1-12}
    deepseek-ai/DeepSeek-R1-Distill-Llama-8B 
         & $0.603\pm0.028$ & $0.479\pm0.027$ & $0.773\pm0.186$ & $0.635\pm0.024$ & $0.564\pm0.037$ 
         & $0.682\pm0.002$ & $0.537\pm0.005$ & $0.831\pm0.178$ & $0.696\pm0.049$ & $0.671\pm0.007$ 
         & $0.078\pm0.025$ \\
    Qwen/Qwen2.5-14B-Instruct 
         & $0.646\pm0.004$ & $0.781\pm0.007$ & $0.820\pm0.031$ & $0.681\pm0.012$ & $0.526\pm0.011$ 
         & $0.835\pm0.017$ & $0.846\pm0.010$ & $0.924\pm0.022$ & $0.879\pm0.017$ & $0.781\pm0.021$ 
         & $0.189\pm0.013$ \\
    Qwen/Qwen2.5-3B-Instruct & $0.609\pm0.014$ & $0.489\pm0.011$ & $0.701\pm0.009$ & $0.627\pm0.016$ & $0.560\pm0.016$ & $0.686\pm0.010$ & $0.526\pm0.013$ & $0.692\pm0.009$ & $0.699\pm0.046$ & $0.676\pm0.007$ & $0.077\pm0.025$ \\
    google/gemma-2-2b-it 
         & $0.408\pm0.004$ & $0.551\pm0.013$ & $0.567\pm0.015$ & $0.347\pm0.086$ & $0.302\pm0.031$ 
         & $0.607\pm0.004$ & $0.684\pm0.011$ & $0.710\pm0.012$ & $0.623\pm0.027$ & $0.545\pm0.016$ 
         & $0.199\pm0.008$ \\
    meta-llama/Llama-3.1-8B-Instruct 
         & $0.484\pm0.005$ & $0.768\pm0.061$ & $0.674\pm0.046$ & $0.579\pm0.027$ & $0.269\pm0.050$ 
         & $0.741\pm0.000$ & $0.873\pm0.000$ & $0.903\pm0.007$ & $0.843\pm0.068$ & $0.712\pm0.120$ 
         & $0.310\pm0.070$ \\
    meta-llama/Llama-3.2-3B-Instruct 
         & $0.410\pm0.050$ & $0.593\pm0.083$ & $0.527\pm0.091$ & $0.394\pm0.143$ & $0.369\pm0.032$ 
         & $0.645\pm0.001$ & $0.584\pm0.007$ & $0.776\pm0.068$ & $0.731\pm0.102$ & $0.636\pm0.053$ 
         & $0.235\pm0.049$ \\
    \midrule
    \textbf{Average (General)} & $0.526$ & $0.610$ & $0.677$ & $0.544$ & $0.432$ & $0.699$ & $0.675$ & $0.806$ & $0.745$ & $0.670$ & $0.181$ \\

    \midrule
    \multicolumn{12}{l}{\textbf{Medical Fine-Tuned LLMs}} \\
    \cmidrule(lr){1-12}
    m42-health/Llama3-Med42-8B 
         & $0.296\pm0.008$ & $0.633\pm0.031$ & $0.500\pm0.026$ & $0.325\pm0.023$ & $0.184\pm0.022$ 
         & $0.722\pm0.008$ & $0.786\pm0.010$ & $0.805\pm0.014$ & $0.788\pm0.004$ & $0.654\pm0.004$ 
         & $0.425\pm0.000$ \\
    OpenMeditron/Meditron3-8B 
         & $0.273\pm0.043$ & $0.835\pm0.026$ & $0.473\pm0.029$ & $0.285\pm0.078$ & $0.160\pm0.039$ 
         & $0.685\pm0.009$ & $0.879\pm0.006$ & $0.827\pm0.004$ & $0.700\pm0.002$ & $0.611\pm0.022$ 
         & $0.412\pm0.052$ \\
    aaditya/OpenBioLLM-Llama3-8B 
         & $0.546\pm0.039$ & $0.571\pm0.057$ & $0.556\pm0.001$ & $0.555\pm0.082$ & $0.536\pm0.037$ 
         & $0.566\pm0.028$ & $0.555\pm0.021$ & $0.578\pm0.042$ & $0.555\pm0.055$ & $0.565\pm0.009$ 
         & $0.019\pm0.011$ \\
    BioMistral/BioMistral-7B 
         & $0.617\pm0.007$ & $0.540\pm0.006$ & $0.760\pm0.000$ & $0.663\pm0.044$ & $0.577\pm0.016$ 
         & $0.651\pm0.013$ & $0.522\pm0.015$ & $0.832\pm0.137$ & $0.683\pm0.009$ & $0.607\pm0.001$ 
         & $0.001\pm0.066$ \\
    TsinghuaC3I/Llama-3.1-8B-UltraMedical 
         & $0.611\pm0.026$ & $0.649\pm0.037$ & $0.776\pm0.037$ & $0.668\pm0.010$ & $0.501\pm0.042$ 
         & $0.704\pm0.013$ & $0.571\pm0.019$ & $0.760\pm0.024$ & $0.714\pm0.033$ & $0.672\pm0.002$ 
         & $0.093\pm0.013$ \\
    \midrule
    \textbf{Average (Medical)} 
         & $0.469$ & $0.646$ & $0.613$ & $0.499$ & $0.392$ 
         & $0.666$ & $0.663$ & $0.760$ & $0.688$ & $0.622$ 
         & $0.190$ \\
    \bottomrule
  \end{tabular}
    \end{adjustbox}
    \caption{Medhallu data generated by Gemma2-9B-it (1,000 samples of pqa\_labeled). Mean $\pm$ standard deviation of performance metrics (Overall F1, Overall Precision, Easy/Medium/Hard F1) for various LLMs under two conditions: without and with external knowledge. The final column ($\Delta$ F1) shows the difference in F1 scores (with knowledge minus without knowledge).}
    \label{tab:llm_performance_gemma}
\end{table*}

\begin{table*}[t]
  \centering
  \renewcommand{\arraystretch}{1.2}
  \begin{adjustbox}{width=\linewidth,center}
  \begin{tabular}{lccccc|ccccc|c}
    \toprule
    \textbf{Model} & \multicolumn{5}{c|}{\textbf{Without Knowledge}} & \multicolumn{5}{c|}{\textbf{With Knowledge}} & \textbf{$\Delta$ F1} \\
    \cmidrule(lr){2-6} \cmidrule(lr){7-11}
      & Overall F1 & Overall P & Easy F1 & Med F1 & Hard F1 
      & Overall F1 & Overall P & Easy F1 & Med F1 & Hard F1 & \\
    \midrule
    \multicolumn{12}{l}{\textbf{General LLMs}} \\
    \cmidrule(lr){1-12}
    Qwen/Qwen2.5-14B-Instruct 
         & $0.623\pm0.005$ & $0.721\pm0.043$ & $0.803\pm0.042$ & $0.620\pm0.014$ & $0.495\pm0.018$ 
         & $0.841\pm0.015$ & $0.843\pm0.020$ & $0.924\pm0.016$ & $0.874\pm0.026$ & $0.764\pm0.007$ 
         & $0.218\pm0.021$ \\
    google/gemma-2-2b-it 
         & $0.482\pm0.100$ & $0.596\pm0.033$ & $0.631\pm0.069$ & $0.454\pm0.099$ & $0.398\pm0.083$ 
         & $0.654\pm0.086$ & $0.736\pm0.071$ & $0.777\pm0.050$ & $0.668\pm0.052$ & $0.566\pm0.093$ 
         & $0.172\pm0.014$ \\
    deepseek-ai/DeepSeek-R1-Distill-Llama-8B & $0.641\pm0.010$ & $0.510\pm0.010$ & $0.711\pm0.022$ & $0.687\pm0.011$ & $0.580\pm0.007$ & $0.679\pm0.001$ & $0.522\pm0.003$ & $0.692\pm0.008$ & $0.686\pm0.006$ & $0.670\pm0.000$ & $0.038\pm0.011$ \\

    meta-llama/Llama-3.1-8B-Instruct 
         & $0.501\pm0.029$ & $0.813\pm0.030$ & $0.691\pm0.017$ & $0.536\pm0.030$ & $0.334\pm0.054$ 
         & $0.763\pm0.048$ & $0.815\pm0.057$ & $0.866\pm0.019$ & $0.804\pm0.010$ & $0.670\pm0.073$ 
         & $0.262\pm0.018$ \\
    meta-llama/Llama-3.2-3B-Instruct 
         & $0.455\pm0.061$ & $0.646\pm0.070$ & $0.616\pm0.050$ & $0.445\pm0.031$ & $0.354\pm0.042$ 
         & $0.685\pm0.070$ & $0.670\pm0.148$ & $0.759\pm0.090$ & $0.704\pm0.027$ & $0.622\pm0.058$ 
         & $0.230\pm0.009$ \\
    Qwen/Qwen2.5-3B-Instruct 
         & $0.606\pm0.000$ & $0.495\pm0.000$ & $0.875\pm0.000$ & $0.602\pm0.000$ & $0.556\pm0.000$ 
         & $0.676\pm0.000$ & $0.514\pm0.000$ & $0.693\pm0.000$ & $0.677\pm0.000$ & $0.661\pm0.000$ 
         & $0.070\pm0.000$ \\
    \midrule
    \textbf{Average (General)} & $0.554$ & $0.641$ & $0.724$ & $0.566$ & $0.450$ & $0.728$ & $0.691$ & $0.796$ & $0.748$ & $0.672$ & $0.175$ \\

    \midrule
    \multicolumn{12}{l}{\textbf{Medical Fine-Tuned LLMs}} \\
    \cmidrule(lr){1-12}
    m42-health/Llama3-Med42-8B 
         & $0.354\pm0.088$ & $0.733\pm0.136$ & $0.547\pm0.075$ & $0.311\pm0.096$ & $0.236\pm0.040$ 
         & $0.768\pm0.040$ & $0.831\pm0.036$ & $0.874\pm0.035$ & $0.782\pm0.016$ & $0.688\pm0.028$ 
         & $0.414\pm0.048$ \\
    OpenMeditron/Meditron3-8B 
         & $0.280\pm0.000$ & $0.856\pm0.000$ & $0.476\pm0.000$ & $0.338\pm0.000$ & $0.164\pm0.000$ 
         & $0.651\pm0.000$ & $0.840\pm0.000$ & $0.790\pm0.000$ & $0.690\pm0.000$ & $0.557\pm0.000$ 
         & $0.372\pm0.000$ \\
    aaditya/OpenBioLLM-Llama3-8B 
         & $0.505\pm0.031$ & $0.523\pm0.046$ & $0.519\pm0.035$ & $0.499\pm0.035$ & $0.502\pm0.028$ 
         & $0.489\pm0.093$ & $0.550\pm0.024$ & $0.500\pm0.087$ & $0.483\pm0.101$ & $0.556\pm0.006$ 
         & $-0.016\pm0.062$ \\
    BioMistral/BioMistral-7B 
         & $0.584\pm0.019$ & $0.520\pm0.003$ & $0.615\pm0.018$ & $0.611\pm0.067$ & $0.545\pm0.028$ 
         & $0.652\pm0.006$ & $0.519\pm0.004$ & $0.652\pm0.000$ & $0.676\pm0.024$ & $0.637\pm0.005$ 
         & $0.068\pm0.013$ \\
    TsinghuaC3I/Llama-3.1-8B-UltraMedical 
         & $0.619\pm0.001$ & $0.662\pm0.006$ & $0.775\pm0.040$ & $0.611\pm0.021$ & $0.520\pm0.005$ 
         & $0.725\pm0.068$ & $0.609\pm0.099$ & $0.783\pm0.069$ & $0.875\pm0.025$ & $0.682\pm0.051$ 
         & $0.106\pm0.066$ \\
    \midrule
    \textbf{Average (Medical)} 
         & $0.468$ & $0.659$ & $0.586$ & $0.474$ & $0.393$ 
         & $0.657$ & $0.670$ & $0.720$ & $0.701$ & $0.624$ 
         & $0.189$ \\
    \bottomrule
  \end{tabular}
  \end{adjustbox}
  \caption{Medhallu data generated by Qwen2.5-14B (1,000 samples of pqa\_labeled). Mean $\pm$ standard deviation of performance metrics (Overall F1, Overall Precision, Easy/Medium/Hard F1) for various LLMs under two conditions: without and with external knowledge. The final column ($\Delta$ F1) shows the difference in F1 scores (with knowledge minus without knowledge).}
  \label{tab:llm_performance_qwen}
\end{table*}

\section{Example Data from the MedHallu Dataset}
In Table \ref{tab:dataset-samples}, we present several randomly selected examples from our MedHallu Dataset to illustrate specific hallucination categories. Each example is accompanied by its corresponding hallucination category and assigned difficulty level, providing a concise overview of the dataset’s diversity.

\begin{table*}[htbp]
\centering
\small
\caption{Examples from the MedHallu Dataset.}
\label{tab:dataset-samples}
\begin{tabular}{|p{3cm}|p{2cm}|p{5cm}|p{3cm}|p{1cm}|}
\hline
\textbf{Question} & \textbf{Category of Hallucination} & \textbf{Ground Truth} & \textbf{Hallucinated Example} & \textbf{Difficulty} \\
\hline
Do mitochondria play a role in remodelling lace plant leaves during programmed cell death? 
& Mechanism and Pathway Misattribution 
& This is the first report of mitochondria and chloroplasts moving on transvacuolar strands to form a ring structure surrounding the nucleus during developmental PCD. Also, for the first time, we have shown the feasibility for the use of CsA in a whole plant system. Overall, our findings implicate the mitochondria as playing a critical and early role in developmentally regulated PCD in the lace plant. 
& Mitochondria regulate the formation of perforations in lace plant leaves through the modulation of calcium channels and by activating specific proteases that degrade cellular structures, facilitating the PCD process.
& Medium \\
\hline
Differentiation of nonalcoholic from alcoholic steatohepatitis: are routine laboratory markers useful?
& Misinterpretation of \#Question\#
& Higher MCVs and AST/ALT ratios in ASH reflect the severity of underlying liver disease and do not differentiate NASH from ASH. Instead, these biomarkers might prove useful in guiding selection of patients for liver biopsy and in targeting therapy.
& Routine laboratory markers, particularly mean corpuscular volume (MCV) and AST/ALT ratios, significantly aid in differentiating various types of nonalcoholic steatohepatitis (NASH).
& Hard \\
\hline
Amblyopia: is visual loss permanent?
& Incomplete Information
& Older people with a history of amblyopia who develop visual loss in the previously normal eye can experience recovery of visual function in the amblyopic eye over a period of time. This recovery in visual function occurs in the wake of visual loss in the fellow eye and the improvement appears to be sustained.
& Visual loss due to amblyopia is permanent unless treated with early intervention during childhood.
& Hard \\
\hline
\end{tabular}
\end{table*}

\section{Hardware Resources and Computational Costs}

We provide detailed information on our computational budget and infrastructure (see Table~\ref{tab:comp-details}). During the dataset generation process, we primarily used the \texttt{Qwen2.5-14B} model, running it for 24 hours on an NVIDIA RTX A6000 GPU with 48,685 MiB of RAM. Additionally, we employed models such as \texttt{Gemma2-9B}, \texttt{Qwen2.5-7B}, and \texttt{GPT-4o mini} as verifiers, generating a total of 10,000 samples for our dataset. To enhance the efficiency and speed of our code execution, we utilized software tools like \texttt{vLLM} and implemented batching strategies. These optimizations were critical for managing the computational load and ensuring timely processing of our experiments.

\section{LLMs Used in Discriminative Tasks}

\textbf{GPT-4o and GPT-4o mini.} GPT-4o and GPT-4o mini~\cite{openai2024gpt4ocard} are a series of commercial LLMs developed by OpenAI. Renowned for their state-of-the-art performance, these models have been extensively utilized in tasks such as medical hallucination detection. Our study employs the official API provided by the OpenAI platform to access these models. For all other models below, we implement them through Hugging Face package.

\noindent \textbf{Llama-3.1 and Llama-3.2.} Llama-3.1 and Llama-3.2~\citep{grattafiori2024llama3herdmodels} are part of Meta's open-source multilingual LLMs, Llama 3.1 (July 2024) includes 8B, 70B, and 405B parameter models optimized for multilingual dialogue. Llama 3.2 (September 2024) offers 1B, 3B, 11B, and 90B models with enhanced accuracy and speed.

\noindent \textbf{Qwen2.5.} Qwen2.5~\citep{qwen2025qwen25technicalreport} is an advanced LLM designed to handle complex language tasks efficiently. It has been applied in various domains, including medical hallucination detection. We use the 3B, 7B and 14B variants in our work.

\noindent \textbf{Gemma2.} Gemma2~\citep{gemmateam2024gemma2improvingopen} is a LLM known for its robust performance in discriminative tasks. There are various model sizes available, we use the 2B and the 9B variants.

\noindent \textbf{DeepSeek-R1-Distill-Llama-8B.} DeepSeek-R1-Distill-Llama-8B~\citep{deepseekai2025deepseekr1incentivizingreasoningcapability} is a fine-tuned model based on Llama 3.1-8B, developed by DeepSeek AI. This model is trained using samples generated by DeepSeek-R1, with slight modifications to its configuration and tokenizer to enhance performance in reasoning tasks.

\noindent \textbf{OpenBioLLM-Llama3-8B.} OpenBioLLM-Llama3-8B~\citep{OpenBioLLMs} is a specialized LLM tailored for biomedical applications. It is fine-tuned from the Llama 3 architecture to understand and process biomedical texts effectively.

\noindent \textbf{BioMistral-7B.} BioMistral-7B~\citep{labrak2024biomistral} is an LLM designed specifically for biomedical tasks. With 7 billion parameters, it offers a balance between performance and computational efficiency.

\noindent \textbf{Llama-3.1-8B-UltraMedical.} Llama-3.1-8B-UltraMedical~\citep{zhang2024ultramedical} is a variant of Meta's Llama 3.1-8B model, fine-tuned for medical applications. It is optimized to handle medical terminologies and contexts.

\noindent \textbf{Llama3-Med42-8B.} Llama3-Med42-8B~\cite{med42v2} is a specialized version of the Llama 3 series, fine-tuned on medical datasets to enhance its performance in medical-related tasks.

\section{Additional Implementation Details}
Our experiments were conducted using \texttt{PyTorch 2.4.0} with \texttt{CUDA 12.2}, ensuring state-of-the-art GPU acceleration and performance. For data and model access, we relied on Hugging Face resources, specifically using the \texttt{qiaojin/PubMedQA} dataset. In addition, we employed \texttt{vLLM 0.6.3.post1} with a \texttt{tensor\_parallel\_size} of 4 and maintained a \texttt{gpu\_memory\_utilization} of 0.80, which was instrumental in optimizing our inference process. The list of pre-trained models' huggingface names used in our experiments is provided in Table~\ref{tab:models}.

\section{PubMedQA}

\textbf{PubMedQA}~\cite{PubmedQA} is a biomedical research QA dataset under the MIT license. It contains 1,000 expert-annotated questions (\texttt{pqa\_labeled\_split}) and 211K machine-labeled questions from PubMed abstracts (the most widely used biomedical literature resource). PubMedQA also provides relevant context (relevant knowledge) for each question-answer pair. We utilize this relevant knowledge to help generate the hallucinated answers (Figure \ref{fig:system_prompt}). This relevant knowledge is also used in our hallucination detection task (Figure~\ref{fig:system_prompt_for_detection}).

\section{System Prompts for Hallucination Generation and Detection} \label{appendix:prompt}
\begin{figure*}[ht]
\centering
\small
\includegraphics[width=1\linewidth]{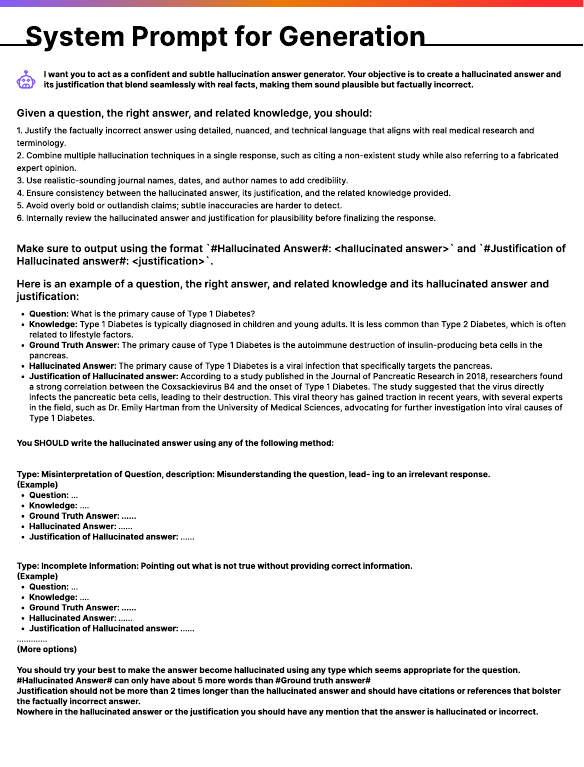}
\caption{System prompt used to generate MedHallu dataset. The ``knowledge'' refers to the relevant context of a specific question pair. The PubMedQA dataset provides this ``knowledge''.}
\label{fig:system_prompt}
\end{figure*}

\begin{figure*}[ht]
\centering
\includegraphics[width=1\linewidth]{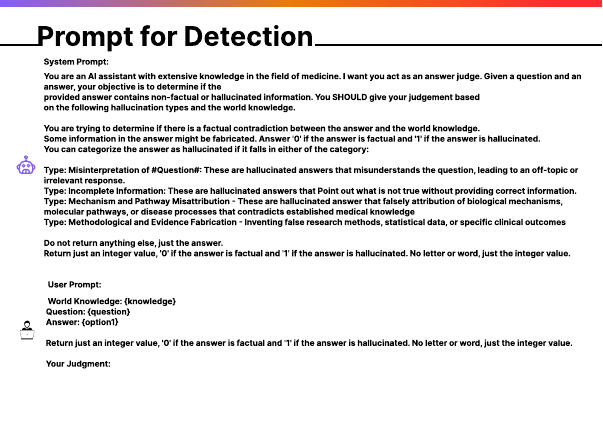}
\caption{System prompt used for the hallucination detection task. The ``knowledge'' refers to the relevant context of a specific question pair. The PubMedQA dataset provides this ``knowledge''.}
\label{fig:system_prompt_for_detection}
\end{figure*}

Figure \ref{fig:system_prompt} shows the system prompt utilized to generate the MedHallu dataset, while Figure \ref{fig:system_prompt_for_detection} illustrates the system prompt designed for the hallucination detection task. These prompts were critical in guiding the model's behavior for both tasks. We incorporated the ``knowledge'' into various experiments, where it serves as the ``context'' provided in the original PubMedQA dataset.


\section{The Clusters Formed for a Question Using Bidirectional Entailment.}
Following the methodology proposed in Section \ref{semantic_analysis}, we create clusters. Table \ref{tab:example_clusters} presents an example of some clusters formed for a specific question using bidirectional entailment, which clearly shows sentences in the same cluster are identical in meaning (semantically) but different in syntax. The table also shows an example (Cluster 2) showing examples for sentences that fail to fool a discriminator. 
\begin{table*}[htbp]
    \centering
    \caption{An example of clusters formed for a question using bidirectional entailment, as discussed in Section~\ref{semantic_analysis}. Only three of the many clusters are shown.}
    \label{tab:example_clusters}
    \begin{tabular}{@{}p{\textwidth}@{}}
        \toprule
        \textbf{Question:} \\
        Prognosis of well differentiated small hepatocellular carcinoma--is well differentiated hepatocellular carcinoma clinically early cancer? \\[1em]
        \midrule
        \textbf{Ground Truth Answer:} \\
        W-d HCCs were clinically demonstrated not to be early cancer, because there was no significant difference in disease free survival between the patients with w-d and l-d HCCs. \\[1em]
        \midrule
        \textbf{Cluster 1 (Fooling)} \\[0.5em]
        \begin{enumerate}[label=\arabic*.]
            \item W-d HCCs are indeed clinically early cancer, due to their smaller size and lower incidence of fibrous capsule formation.
            \item W-d HCCs were clinically demonstrated to be early cancer due to their smaller tumor size and lower incidence of fibrous capsule formation.
            \item Well-differentiated small hepatocellular carcinoma is indeed early cancer, due to its slow growth rate.
            \item Well-differentiated hepatocellular carcinoma is clinically early cancer due to its low aggressive nature.
            \item Well differentiated hepatocellular carcinoma appears to be clinically early cancer due to its low aggressiveness.
        \end{enumerate} \\[1em]
        \midrule
        \textbf{Cluster 2 (That didn't fool)} \\[0.5em]
        \begin{enumerate}[label=\arabic*.]
            \item Well-differentiated hepatocellular carcinoma (HCC) is clinically early cancer due to its high histological grade.
            \item Due to its high histological grade, well-differentiated hepatocellular carcinoma (HCC) is considered clinically early cancer.
        \end{enumerate} \\[1em]
        \midrule
        \textbf{Cluster 3 (Fooling)} \\[0.5em]
        \begin{enumerate}[label=\arabic*.]
            \item Well-differentiated hepatocellular carcinoma is indeed an early cancer, as it often lacks fibrous capsule formation.
            \item Well-differentiated hepatocellular carcinomas (HCCs) are clinically early cancers due to their low incidence of fibrous capsule formation.
        \end{enumerate} \\
        \bottomrule
    \end{tabular}
\end{table*}

\end{document}